\documentclass{sig-alternate-2013}

\newfont{\mycrnotice}{ptmr8t at 7pt}
\newfont{\myconfname}{ptmri8t at 7pt}
\let\crnotice\mycrnotice%
\let\confname\myconfname%

\permission{Permission to make digital or hard copies of all or part of this work for personal or classroom use is granted without fee provided that copies are not made or distributed for profit or commercial advantage and that copies bear this notice and the full citation on the first page. Copyrights for components of this work owned by others than the author(s) must be honored. Abstracting with credit is permitted. To copy otherwise, or republish, to post on servers or to redistribute to lists, requires prior specific permission and/or a fee. Request permissions from permissions@acm.org.}
\conferenceinfo{KDD '13,}{August 11--14, 2013, Chicago, Illinois, USA. \\
{\mycrnotice{Copyright is held by the owner/author(s). Publication rights licensed to ACM.}}}
\CopyrightYear{2013} 
\crdata{978-1-4503-2174-7/13/08} 
\usepackage{cite}
\usepackage{graphicx}
\usepackage{subfig}
\usepackage{url}
\usepackage{xspace}
\usepackage{algorithmic}
\usepackage{algorithm}
\usepackage{multirow}

\newdef{defn}{Definition}

\def\ie{\emph{i.e.,~}}
\def\eg{\emph{e.g.,~}}

\newcommand{\appname}{{\sc KERA}\xspace}

\newcounter{Lcount}
\newcommand{\numsquishlist}{
  \begin{list}{\arabic{Lcount}. }
   { \usecounter{Lcount}
 \setlength{\itemsep}{0pt}      \setlength{\parsep}{3pt}
     \setlength{\topsep}{3pt}       \setlength{\partopsep}{0pt}
     \setlength{\leftmargin}{2.7em} \setlength{\labelwidth}{1em}
     \setlength{\labelsep}{0.5em} } }
\newcommand{\numsquishend}{\end{list}}

\newcommand{\squishlist}{
  \begin{list}{$\bullet$}
   { \setlength{\itemsep}{0pt}      \setlength{\parsep}{3pt}
     \setlength{\topsep}{3pt}       \setlength{\partopsep}{0pt}
     \setlength{\leftmargin}{1.5em} \setlength{\labelwidth}{1em}
     \setlength{\labelsep}{0.5em} } }
\newcommand{\squishend}{\end{list}}

\usepackage{color, colortbl}
\definecolor{Gray}{gray}{0.9}
\definecolor{LightCyan}{rgb}{0.88,1,1}

\begin{document}
%

\title{Exploratory Analysis of \\Highly Heterogeneous Document Collections}

\numberofauthors{1}

\author{
\alignauthor
Arun S. Maiya, John P. Thompson, Francisco Loaiza-Lemos, and Robert M. Rolfe\\
       \affaddr{Institute for Defense Analyses --- Alexandria, VA, USA}\\
       \email{\{amaiya, jpthomps, floaiza, rolfe\}@ida.org}
}


\newcounter{copyrightbox}

\maketitle
\begin{abstract}
We present an effective multifaceted system for exploratory analysis of highly heterogeneous document collections.  Our system is based on intelligently tagging individual documents in a purely automated fashion and exploiting these tags in a powerful faceted browsing framework.  Tagging strategies employed include both unsupervised and supervised approaches based on machine learning and natural language processing.  As one of our key tagging strategies, we introduce the \appname algorithm ({\bf K}eyword {\bf E}xtraction for {\bf R}eports and {\bf A}rticles).  \appname  extracts topic-representative terms from individual documents in a purely unsupervised fashion and is revealed to be significantly more effective than state-of-the-art methods.  Finally, we evaluate our system in its ability to help users locate documents pertaining to military critical technologies buried deep in a large heterogeneous sea of information.

\end{abstract}


%
%

\section{Introduction and Motivation}
\label{sec:intro}


Given a large and diverse collection of unstructured text documents, how does one (1) characterize the subject areas present and (2) use these discovered subject areas to efficiently navigate the collection to locate critical information?  Many previous works have investigated such questions within specific domains such as microblog posts (\eg characterizing tweets \cite{Kumar2012Navigating}), but comparatively less attention has been paid to investigating more general and diverse contexts.  Unfortunately, in practice, approaches that may work well for domains consisting exclusively of a single document type (\eg tweets, emails, or scientific articles) do not always translate easily or directly to other more {\em heterogeneous} and ``messy'' document collections. In this work, we present a tag-based system in which tags (\ie terms or character strings automatically assigned to individual documents) are exploited to efficiently characterize and explore document collections.  Document collections of interest in our work exhibit a high degree of diversity (\eg arbitrary files residing on a high-capacity laptop drive or file server).  The U.S. federal government, for instance, is often presented with the challenge of what essentially is exploratory analyses of highly heterogeneous document collections.  These collections requiring analyses are nowhere near as homogeneous as tweets, news stories, patents, or scientific abstracts --- the typical objects of study in text analytics research (\eg \cite{Rose2010Automatic,Bun2002Topic,Tang2012PatentMiner,Wei2010TIARA,He2009Detecting}).  To better illustrate this point, we briefly describe three motivating examples.

\begin{enumerate}
 \item {\bf Digital Investigations:}   In 2012, it was reported that General David Petraeus, CIA Director, had been having an affair with his biographer, Paula Broadwell \cite{Pearson2012Petraeus}.  In the ensuing investigation, the FBI discovered that a laptop owned by Paula Broadwell may have contained sensitive classified information, constituting a security violation \cite{Pearson2012Petraeus}.  Reviewing computers for sensitive or critical information is a task that arises in many scenarios. Examples include digital forensics and the identification of critical information (\eg trade secrets) exposed through cyber intrusions.  Performing such reviews and making such discoveries can be extremely difficult and burdensome, as analysts are faced with the challenge of locating critical information buried deep in a large heterogeneous sea of files. 

 \item  {\bf Intelligence Analysis:}  Intelligence Analysis generally involves acquiring knowledge of subjects, entities, or situations of interest and characterizing and understanding possible future scenarios \cite{MacEachinTradecraft}.  Much of this situational awareness is achieved through analyses of unstructured text collections comprised of diverse sets of document types and file formats.  The size and breadth of information embedded in such collections can be overwhelming to intelligence analysts.
 \item {\bf Appraisal of Electronic Records:} The National Archives and Records Administration (NARA) is charged with determining the value of federal records for archival purposes \cite{Lee2008Text}.  This process, known as {\em document appraisal}, makes documents either permanent or temporary and involves time-consuming reviews of massive collections of documents that are diverse in both their content and format \cite{Lee2008Text}. 
\end{enumerate}

~\\
All three of the above examples necessitate the need for an efficient and intelligent way to explore heterogeneous, large-scale document collections for critical information of interest.  In the present work, we explore the task of characterizing, browsing, and searching large collections of unstructured text documents using {\em faceted navigation}.  We present a system that effectively discovers and exploits the use of {\em information facets} to efficiently characterize and search large document collections.

\subsection{Information Facets}
\label{sec:intro.facets}
{\em Information facets} (or simply {\em facets}) are classes of attributes describing objects in an information repository.  They are used to facilitate searches, filtering, and navigation of the information by different dimensions of the data \cite{Tunkelang2009Faceted,Ranganathan1962Elements}. Faceted classification systems were first conceived in the 1930s by S.R. Ranganathan, an Indian librarian considered to be the father of library science \cite{Ranganathan1962Elements}.  Today, faceted search is used extensively in information retrieval systems (\eg \cite{Tunkelang2009Faceted}).\footnote{Faceted classification systems, in the context of computer systems, are also referred to as {\em faceted search}, {\em faceted navigation}, or {\em faceted browsing}.}  Electronic commerce sites, for instance, employ facets to facilitate browsing products along various dimensions (\eg brand, price).  The vast majority of faceted search systems populate the facet attributes from {\em pre-existing} fields in a data repository.  One example is Twitter's use of {\em hashtags}, which are user-generated topic tags assigned to tweets \cite{Kumar2012Navigating}.  Another such example is an {\em author} or {\em title} field in a publication database.  Unfortunately, for most large document collections, these manually-generated tags typically do not exist.  For these cases, the attributes used to populate facets must be mined or {\em discovered}.

What facets should be used for unstructured text and how might we populate them in an automated fashion (\ie discover them)?  In the context of a physical library system, Ranganathan proposed classifying information according to five, manually-populated facet categories he referred to as Personality, Matter, Energy, Space, and Time (PMEST) \cite{Ranganathan1962Elements}. Motivated by the PMEST model and today's surge in electronic records, we propose organizing unstructured digital text collections by the following general set of discoverable facet categories.\footnote{By ``discoverable'', we mean that the attributes for these facets are populated in an automated fashion.}

~\\
\noindent
{\bf Topic Facets.}  Topic Facets relate to the overall subject or ``aboutness'' of an electronic document.\footnote{Sentiment analysis might also be included here.}  In the present work, our focus is on the automated discovery of {\em topic tags}.  {\em Topic tags} are key terms that capture or represent the overall topic of the document.   Such tags can be used to characterize and navigate document collections and refine search results.  The problem of discovering topics (and topic tags) from text collections has been extensively studied, of course.  In Section \ref{sec:topic}, we discuss multiple approaches to building {\em Topic Facets}.

~\\
\noindent
{\bf Mention Facets.}  Documents often make mention of entities, relations, or events that are of importance despite being unrelated to the overall topic of the document. For instance, Personally Identifiable  Information (PII) such as social security numbers are often important to detect, yet are not necessarily representative of the subject of a given document.  The same is true of username mentions in tweets.  Mention Facets are populated by extracting such entities (or relations) from text and can be used to help discover and locate information of interest within a document collection.

~\\
\noindent
{\bf Format Facets.}  The Format Facet allows navigation of document collections by file type and format.  For homogeneous text collections like tweets, the Format Facet is rather uninteresting, as all tweets have the same format (\eg plain text and up to 140 characters) .  In our work, however, the document sets from which topic tags and entity mentions are extracted are highly diverse in both type and format.  Collections can include technical reports, news articles, Powerpoint briefs, Excel files, Web pages, programming language source code, emails, and many other files.  In practice, the availability of a Format Facet is especially important for helping to hone in on particular information elements of interest.  For instance, different file types often cover different sets of topic areas and entities.  Moreover, the file type and format affect the way in which topics, entities, and terms of interest are extracted (\eg  a scientific report in PDF format {\em vs.} a .dat file containing Web search history).  
    
~\\
\noindent
{\bf Location Facets.} The meaning and implementation of the Location Facet is subject to interpretation and choice depending on the application.  For documents residing on a file server or workstation, the Location Facet may comprise file paths or folders of documents.  For tweets, on the other hand, it would make most sense to use the geolocation information of the tweet for this facet.

~\\
\noindent
{\bf Time Facets.}  The way in which the Time Facet is implemented will also depend on the application.  For tweets, it might be the time the tweet was posted, whereas an extracted publication date would be of most interest for scientific articles.  For arbitrary files residing on a file server or laptop drive, it is sometimes useful to utilize the {\em created} or {\em last-modified} date for the Time Facet (which must be extracted from the file metadata).

~\\
\noindent
{\bf Author Facets.} As with the Location and Time facets, use of the Author facet will vary by application domain.  In the case of news articles, the author might be extracted from the document content.  For tweets, it will simply be the Twitter user account producing the tweet.  For documents produced by Microsoft Office applications (\eg Microsoft Word, Excel, and Powerpoint), the author might be taken as the {\em Last-Author} field extracted from document metadata.
~\\

Throughout the remainder of this paper, we describe a concrete implementation of our proposed faceted classification system.  First, we summarize our contributions. 


\subsection{Contributions}
\label{sec:intro.contributions}

In this work, we explore the general problem of characterizing, navigating, and searching large collections of diverse documents.  Our contributions are as follows:

\begin{itemize}
 \item  Based on our proposed faceted classification system, we present a fully implemented application for exploring highly heterogeneous document collections that span a wide array of file types, formats, and subject areas.  The tool is designed to facilitate the identification of documents pertaining to military critical technologies, but can readily be used as a general-purpose tool for exploratory analyses of arbitrary text collections.   Our system is based on {\em intelligently tagging} individual documents in a largely automated fashion.
 \item  We propose the \appname  ({\bf K}eyword {\bf E}xtraction for {\bf R}eports and {\bf A}rticles) algorithm, as one means by which informative {\em Topic Facets} can be constructed in an unsupervised and automated fashion.
 \item  For scenarios where \appname may be inappropriate, we present supplemental strategies to characterize and locate information of interest based on supervised machine learning (\ie LinearSVM), unsupervised machine learning (\ie latent Dirichlet allocation or LDA), and natural language processing (\eg Named Entity Recognition or NER).
 \item  We evaluate the application using two separate case studies at sites of deployment.
 \end{itemize}
~\\
\noindent
We begin a discussion of our work by providing an overview of our implemented system.

\section{Application Overview}
\label{sec:appoverview}

Sponsors of our research were in need of a tool to analyze arbitrary document collections in order to help analysts identify documents pertaining to military critical technologies.  The term ``military critical'' here is defined by senior officials and subject matter experts.  (We will sometimes simply use the term ``critical'' when referring to such technologies.)   However, over time, they became interested in the general problem of characterizing, browsing, and searching through arbitrary and heterogeneous document collections (where the definition of critical will vary by application and even user).  The document collections of interest here contain diverse sets of file types and formats that typically exist on file server and workstation drives.  Examples include PDF articles and reports, Microsoft Office documents, plain text log files containing Web browser history, HTML documents, programming language source code files, and more.   Our application is the end result of these objectives and interests.

\begin{figure*}[htb]
\begin{center}
\centerline{\includegraphics[scale=0.38]{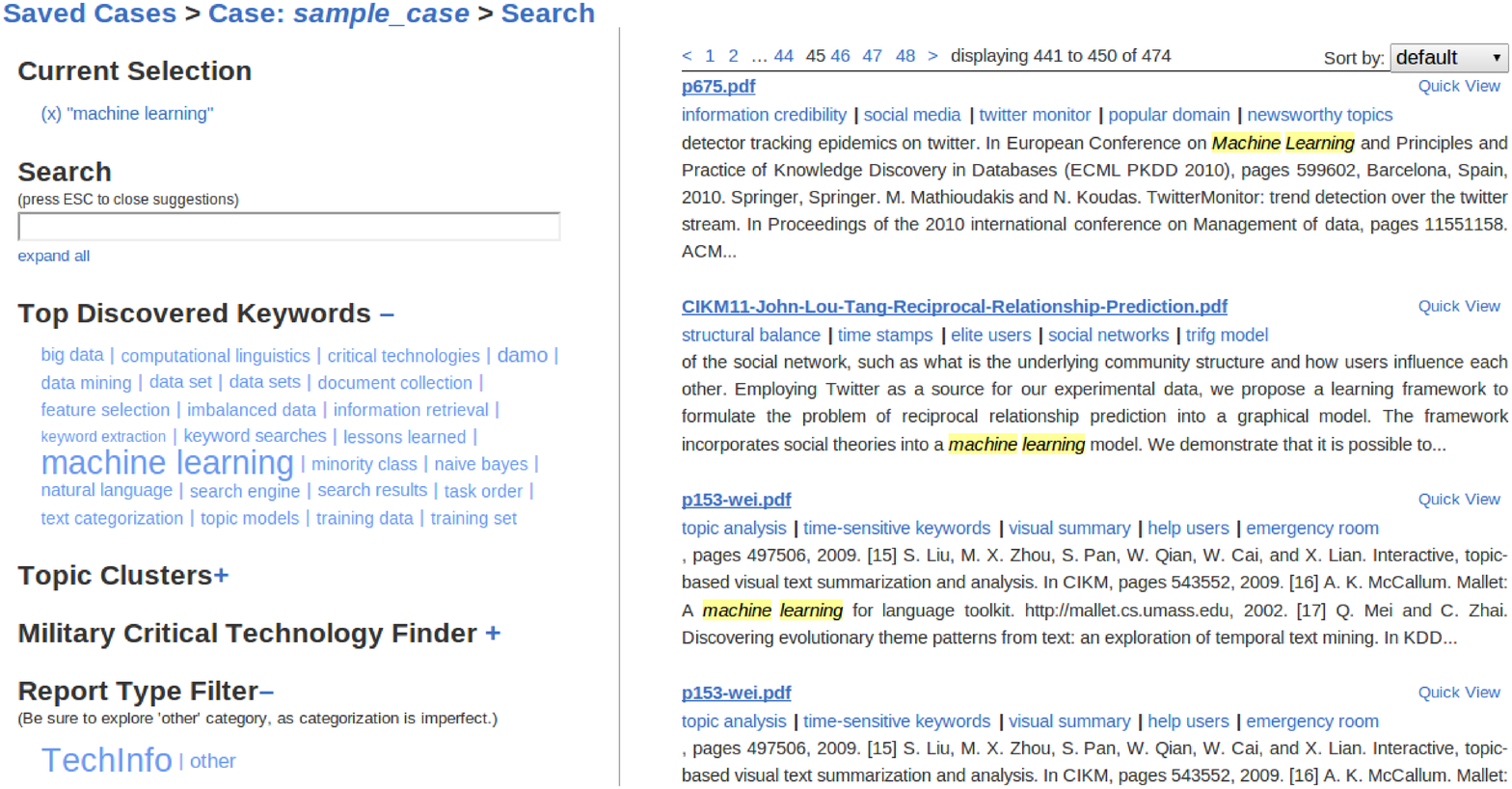}}
\caption{{\scriptsize A screenshot of our system.  On the left, a rich set of information facets are provided for exploratory analysis. Only a subset (\ie the {\em Topic Facets}) is viewable in this screenshot.  The application also provides standard search engine functionality powered by Solr, as shown.}}
\label{fig:appscreenshot}
\end{center}
\vskip -0.3in
\end{figure*}
Figure \ref{fig:appscreenshot} shows a screenshot of one of the main interfaces.  On the surface, it appears to be a standard search engine interface where users can type {\em ad hoc} search queries and view search results.  However, the standard search functionality is enhanced (on the left in Figure \ref{fig:appscreenshot}) with numerous information facets based on the faceted classification system described in Section \ref{sec:intro.facets}.  The facets are populated by intelligently tagging each document in the collection along various dimensions. Documents can be viewed either in their original form or using a ``Quick View'' feature in which case the plain text is shown with highlighted terms (\eg discovered topic-representative keywords, search terms entered by user).  Most (but not all) of the facets take the form of tag clouds.  A {\em tag cloud} is a visualization of a set of words where the relative sizes of the words are determined by either features of the word or features of the entity represented by the word.  In our work, the sizes of tags indicate the number of documents assigned the tag.  Tag clouds are used both as a visualization and as an interface for faceted browsing of document collections, as the tags in the cloud can be used to filter and refine search results.   Each facet can be expanded or collapsed by clicking the $+$ and $-$ symbols, as shown in Figure \ref{fig:appscreenshot}.  The four facets viewable in Figure \ref{fig:appscreenshot} (\eg {\em Top Discovered Keywords}, {\em Topic Clusters}) all fall under the category of {\em Topic Facet} from our aforementioned faceted classification system.  For the remaining facets, there is a one-to-one correspondence with the facet types described in Section \ref{sec:intro.facets}.  All facets (both those appearing in Figure \ref{fig:appscreenshot} and not) are described at length later.

\begin{figure}[htb]
\begin{center}
\centerline{\includegraphics[scale=0.23]{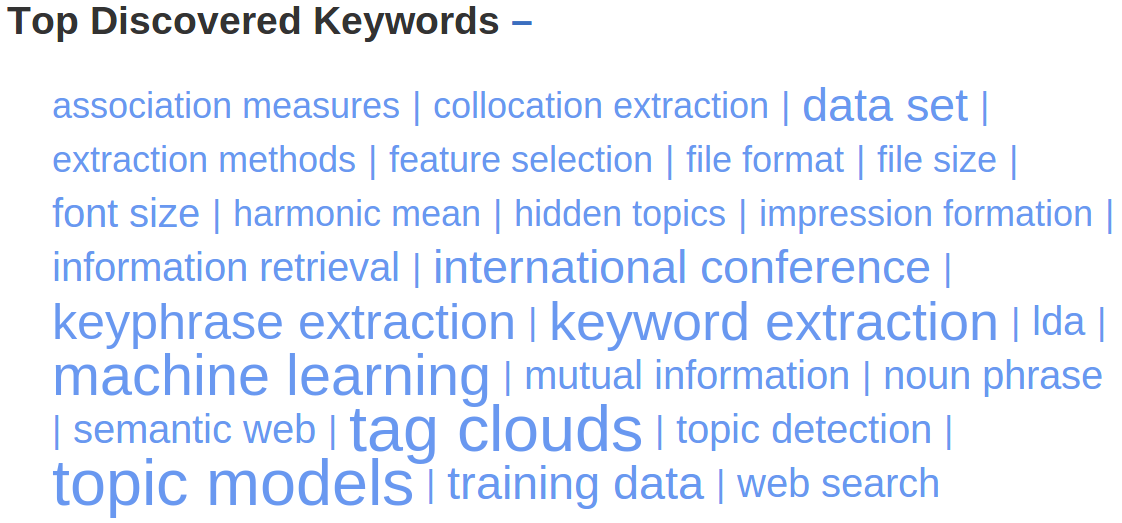}}
\caption{{\scriptsize A tag cloud generated by the \appname algorithm for 64 documents used as references in this paper. (Only a subset were cited due to space constraints.) \appname is discussed in Section \ref{sec.topic.tags}.}}
\label{fig:topictagfacet}
\end{center}
\vskip -0.4in
\end{figure}

~\\
\noindent
{\bf  Tag Clouds as Lenses.}  Figure \ref{fig:topictagfacet} shows a sample tag cloud displaying topic-representative keywords discovered using \appname, our unsupervised algorithm for keyterm extraction.  This tag cloud facet can be viewed as a ``lens'' into document collections.  The remaining facets may be viewed as controls used to point, zoom, and focus this ``lens'' to areas of high interest in the corpus.\footnote{In actuality, each facet can play either the role of a ``lens'' or a ``lens control.''}  For instance, when filtering the search results by folder using a {\em Location Facet}, the tag cloud shown in Figure \ref{fig:topictagfacet} will dynamically re-generate to display the top discovered keywords of only the refined search results (\ie documents residing in the folder selected).  In this way, users can quickly ``triage'' noisy document collections for information of interest (in some cases {\em even before opening and reading documents}).   Although tag clouds have come under criticism in the past, our tag-based system is demonstrated to be surprisingly effective in locating critical information of interest buried deep within document collections.   The key to achieving this success is constructing informative clouds free from noise.  Throughout later sections, we describe how precisely we accomplish this.  But first, we briefly describe the implementation of our system.

~\\
\noindent
{\bf  Implementation Details.}  The underlying engine driving our application is the Solr search server,\footnote{\url{http://lucene.apache.org/solr/}} which natively supports text extraction, full text search, and faceted navigation.    Documents in the Solr index are tagged using a series of supervised and unsupervised data mining algorithms, and it is these tags that power faceted browsing and tag clouds.  All data mining algorithms are developed using the Python language and libraries including scikit-learn,\footnote{\url{http://scikit-learn.org/stable/}} NLTK,\footnote{\url{http://code.google.com/p/nltk/}} and Gensim topic modeling toolkit.\footnote{\url{http://radimrehurek.com/gensim/}} Moreover, all algorithms are implemented to process documents in a stream using both online and parallel processing.  The graphical user interface is implemented using Flask\footnote{\url{http://flask.pocoo.org/}} and AJAX-Solr.\footnote{\url{http://github.com/evolvingweb/ajax-solr/wiki}}  Finally, communication between Python scripts and Solr is handled using the pysolr library.\footnote{\url{http://code.google.com/p/pysolr/}}  For the rest of this paper, we describe the concrete facets employed in our system and the analytics algorithms used to populate them.

\section{Topic Facets}
\label{sec:topic}

{\em Topic Facets} are intended to help discover and characterize the subject areas present in a document collection.  Moreover, they allow users to better navigate the collection to find information of interest.   The first {\em Topic Facet} we discuss is based on unsupervised keyterm extraction.

\subsection{Automated Keyword Extraction}
\label{sec.topic.tags}

Our first approach to populating a {\em Topic Facet} is based on extracting topic-representative terms (\ie keywords) from  documents (shown as {\em Top Discovered Keywords} in Figures \ref{fig:appscreenshot} and \ref{fig:topictagfacet}).  Here, we present the \appname  algorithm ({\bf K}eyword {\bf E}xtraction for {\bf R}eports and {\bf A}rticles). \appname is an {\em unsupervised} algorithm to extract keywords from {\em individual} unstructured text documents (\ie it does not require an entire corpus like TF-IDF and other strategies).  At its core, \appname is a {\em descriptive} model for keyword assignment.  It is based on several key observations of human-assigned keywords\footnote{We refer here to those keywords that {\em also} appear as terms within the document itself.} (especially those in scientific and technical publications):

\squishlist
 \item Many keywords assigned are multi-word terms (as opposed to unigrams).
 \item Most keywords are noun phrases.
 \item When a single-word term is used as a keyword, it often appears in the title (or abstract) or is a {\em proper} noun.  Proper nouns often indicate an algorithm, system, or program being described or employed.
\item If a keyword is composed of three or more terms, it can often (but not always) be split in two and maintain a high level of expressivity of the underlying topic described.  For instance, ``stream data mining'' splits into ``stream data'' and ``data mining,'' and ``social network analysis'' splits into ``social network'' and ``network analysis.''
\squishend

Many of these observations have also been noted in other works (\eg \cite{Rose2010Automatic}).  The \appname algorithm is shown in Algorithm \ref{alg1}.  We now describe its three main components:  collocation extraction, part-of-speech filtering, and ranking.

\begin{algorithm}[htb]
\caption{KERA algorithm}
\label{alg1}
\begin{algorithmic}[1]
\REQUIRE $D$, an unstructured text document
\REQUIRE $K$, the number of keywords to extract
\STATE \# generate candidate keywords
\STATE $terms1=extractCollocations(D)$
\STATE $terms2=extractNounPhrases(D)$
\STATE $terms3=extractProperNounUnigrams(D)$
\STATE $candidates =(terms1\cap terms2)\cup terms3$
\STATE \# rank candidates
\FORALL{$c \in candidates$}
\IF{c is unigram}
\STATE $\alpha = \text{normalized frequency of term } $c$ \text{ in } D$
\ELSE
\STATE $\alpha =$ normalized collocation score
\ENDIF
\STATE $\beta = 1- \frac{\text{index of first occurrence of } c \text{ in }D}{\text{num. of words in D}}$  
\STATE rank score of term $c=\frac{2\cdot \alpha\cdot \beta}{\alpha+\beta}$
\ENDFOR
\STATE \# optionally prune based on domain-specific criteria
\STATE \# $candidates = prune(candidates)$
\STATE return top $K$ candidates based on rank score
\end{algorithmic}
\end{algorithm}
~\\
\noindent
{\bf Collocation Extraction.}  We first employ the use of {\em collocation extraction} to identify candidate key terms.  A collocation is ``an expression consisting of two or more words that corresponds to some conventional way of saying things.''\cite{Manning1999Foundations}  We posit that it is these sets of words that are most likely to contain topic-representative phrases.  Although it is possible to extract collocations of three or more terms, we find that such phrases do not lend themselves to aggregation (\eg for use in tag clouds).  At the same time, we find that one word terms are not expressive enough for users to discern the topics of documents.  Thus, we extract only collocated {\em bigrams} (\ie two-word expressions).  Although  our system supports multiple collocation extraction strategies including the log-likelihood ratio test\cite{Dunning1993Accurate} and Pointwise Mutual Information (PMI) \cite{Manning1999Foundations}, we currently use the log-likelihood ratio exclusively, as we find it performs best with respect to {\em Topic Facets}.  Using the log-likelihood ratio test, the collocation score for a bigram  of words $w_1$ and $w_2$ is $2\sum_{ij}n_{ij}\log\frac{n_{ij}}{m_{ij}},$ where $n_{ij}$ are the observed frequencies of the bigram from the contingency table for $w_1$ and $w_2$ and $m_{ij}$ are the expected frequencies assuming that the bigram is independent\cite{Manning1999Foundations,Dunning1993Accurate}.

~\\
\noindent
{\bf Part-of-Speech Filtering.}  As mentioned, the most expressive keywords are typically noun phrases.  Thus, we filter the set of collocations by removing terms that do not match the pattern {\sc (adjective)*(noun)+}.  If the extracted phrases are greater than two terms, we begin truncating from the left until we are left with a bigram.  Such filtering also helps to remove bogus words sometimes introduced by the text extraction process for non-plain-text document formats.  To this filtered set, we {\em add} extracted unigrams that are proper nouns, as we find such terms can be critical to the topic of documents.  This is especially true of government, scientific, and technical publications, as proper nouns often refer to a system, algorithm, program, or initiative being described.

~\\
\noindent
{\bf Ranking Keywords.}  Finally, we rank the extracted terms, as shown in Algorithm \ref{alg1} and return the top $K$ candidates.\footnote{Currently, we set $K=5$ or $K=10$ for \appname.}  Our ranking methodology takes into account both the position of terms within a document and the collocation score (or term frequency).  The final score is taken as the harmonic mean of these metrics.  Prior to returning the final set, one might optionally prune the candidates based on domain-specific criteria.  For instance, in our case, the set of proper noun unigrams may be pruned to only contain those unigrams that are upper-case, since it is those terms that often signify important technical systems and programs.\footnote{Other possible variations include discarding candidates when proper noun unigrams also appear as part of extracted bigrams, removal of unigrams that do not first appear until later in the document, significance testing to filter the set of collocations, and setting $\alpha$ always as normalized frequency.}


~\\
\noindent
{\bf Comparison to Other Approaches.}  Development of \appname was motivated by the fact that existing algorithms did not meet one or more of our needs.  For instance, a number of the existing approaches are either supervised, require an entire corpus, or both.  Such characteristics are unacceptable, as supervised approaches are labor-intensive and corpus-based methods (such as those like TF-IDF that use inverse document frequency) may undervalue terms associated with prevalent topics. TextRank \cite{Mihalcea2004TextRank} and RAKE \cite{Rose2010Automatic} are two methods that are both unsupervised and operate on {\em individual} documents.  Unfortunately, although both methods can perform reasonably well when supplied only paper abstracts, they sometimes perform less well on longer, messier, and more realistic document structures.\footnote{We base these statements on the TextRank implementation available on \url{http://cpan.org} (\ie Text-Categorize-Textrank-0.51) and the RAKE implementation at \url{http://github.com/aneesha/RAKE}. Full experimental results are omitted due to space limitations.}  For instance, Table  \ref{tab:keracompare} shows the keywords extracted for this very paper.  Note that the keywords extracted by \appname are qualitatively superior to TextRank \cite{Mihalcea2004TextRank} and RAKE \cite{Rose2010Automatic}. 

\begin{table}[thb]
\centering
{\scriptsize
\begin{tabular}{|l|l|} \hline
{\bf Method} &  {\bf Top 5 Extracted Keywords for This Paper}\\ \hline\hline
\multirow{2}{*}{\appname} &  ``document collections'', ``KERA'', ``tag clouds'', \\
& ``machine learning'', ``LDA'' \\ \hline
\multirow{3}{*}{RAKE} & ``plain text log files containing web browser history'',  \\
& ``files spanned numerous file formats including MS office'', \\ 
& ``dat file containing web search history'',\\
& ``provides standard search engine functionality powered'',\\
& ``faceted browsing framework yields significant advantages''\\ \hline
\multirow{2}{*}{TextRank} &  ``document collections'', ``system'', ``document'', \\
& ``tagging'', ``faceted''  \\ \hline
\end{tabular}
\caption{{\scriptsize Top 5 extracted keywords for this paper ordered by rank assigned by each method.  \appname-generated terms are most expressive and well-suited to tag cloud aggregation.  TextRank is also too slow for our needs.}}
\label{tab:keracompare}
}
\vskip -0.1in
\end{table}

\subsection{Topic Modeling and Clustering}
\label{sec:topic.lda}
A second {\em Topic Facet} we employ is based on the concept of topic clusters.  Topic modeling and clustering algorithms segment documents into groups, where the intent is for each group to consist of documents pertaining to a particular topic or theme.  Whereas many clustering algorithms produce ``hard'' clusters or disjoint sets of documents, topic models produce ``soft'' or overlapping clusters.  Topic models and clustering strategies may also tag clusters with topic-representative words.  In topic models like latent Dirichlet allocation or LDA \cite{Blei2003Latent}, topics are modeled as word probability distributions, and these tags are simply the most probable words in a distribution.   Our application supports {\em multiple} approaches to topic clustering including LDA \cite{Blei2003Latent}, Hierarchical Dirichlet Process (HDP) \cite{Teh2006Hierarchical}, Latent Semantic Indexing (LSI) \cite{Manning2008Introduction}, and K-Means \cite{Hastie2003Elements}.  All approaches are provided by the machine learning libraries mentioned in Section \ref{sec:appoverview}.  For the current deployment, we employ LDA exclusively.  Documents are assigned to a topic only if the topic proportion assigned by LDA is greater than $0.3$, and documents are tagged using the top $10$ LDA-derived topic tags.  LDA requires the number of topics, $K$, as input, and we currently set this heuristically based on the size of the document collection.  However, in the future, we plan to migrate to HDP, which is a non-parametric approach to topic modeling \cite{Teh2006Hierarchical}.  The facet populated by LDA is labeled ``Topic Clusters'' and appears as a menu showing the list of discovered topics.  These topic clusters are labeled by LDA-derived tags and ordered by the topic ranking methodology described in \cite{Wei2010TIARA}.

\subsection{Document Classifier Facets}
\label{sec:topic.classifier}
All {\em Topic Facets} discussed  thus far (including topic models) are focused on identifying trends and hotspots within the topic collection.  That is, they are not well-suited to finding ``needles in haystacks.''  A document pertaining to a lone topic of high interest to a particular user may not be identifiable in the presence of large topic clusters displayed in a tag cloud or other interface.   To address this, we supplement the facets populated by \appname and LDA with additional tag cloud facets populated with {\em supervised} document classification.   We have previously reported our work on supervised machine learning for critical technologies in \cite{Maiya2012Supervised} .  Thus, we only include brief and sparse descriptions here.  For more information on the development of document classifiers in this domain, please see \cite{Maiya2012Supervised}.

\subsubsection{Military Critical Technology Finder}
\label{sec:topic.classifier.mct}
The facet labeled {\bf Military Critical Technology Finder} in Figure \ref{fig:appscreenshot} is populated using a set of binary supervised machine learning classifiers.  Each binary classifier is trained to identify documents pertaining to a particular critical technology, and each tag in the cloud represents the positive class of a classifier.  For any individual document, if no binary classifier categorizes the document as positive, then the document is assigned the tag ``other'', which also appears in the cloud.   We use LinearSVM as our main learning algorithm for all classifiers.    Constructing training sets for these classifiers poses a number of challenges.  For instance, when training these binary classifiers for arbitrary file collections (\eg a workstation hard drive), the negative class becomes {\em highly} heterogeneous.  If this heterogeneity is not represented or otherwise addressed in the training set, performance can degrade.  In addition, documents pertaining
to critical technologies can sometimes compromise a very small
minority of all possible files encountered.  This is known as the class imbalance problem and can also cause performance to suffer due to bias.  To address these and other problems, we employ heavy use of {\em active learning} in a two step sampling procedure \cite{Maiya2012Supervised}.  We first employ active learning strategies (\eg minimum marginal hyperplane) to sample only the most informative of negative examples for the initial training set (which helps address heterogeneity).  We, then, balance the training set by further sub-sampling this initial training set to produce the final training set.  

\subsubsection{Report Type Filter}
\label{sec:topic.classifier.reporttype}
Using a very similar methodology to the one described in the previous section, we develop an additional classifier to categorize documents based on report type.  That is, documents are categorized into one of four categories:   Technical Information (\eg a research paper), Test Information (\eg a test plan for a system), Programmatic Information (\eg details of a program for  development of a system), and Other (\ie everything else).

\section{Mention Facets}
\label{sec:mention}
Users sometimes may be interested in locating documents not by topic but by mentions of particular entities, terms, or expressions of interest (\eg IP addresses).  To address this, we employ the use of a {\em Mention Facet}, which allows users to upload a plain text file containing expressions of interest.  These expressions can currently take the form of simple lists of terms, gazetteers (\ie entity dictionaries), or regular expressions for patterns of interest (\eg a social security number).  The results are displayed as either a tag cloud or menu, where the items are either explicit terms with matches in the document collection or high-level categories described by expressions (\eg tagging documents containing social security numbers with ``PII'').  Recall that our current research sponsor is specifically interested identifying military critical information.  Thus, for the first deployment of our application, we populate the {\em Mention Facet} in the following manner.  We take the training sets used for our binary classifiers described in Section \ref{sec:topic.classifier.mct} and extract the top $25$ most discriminative terms based on {\em information gain} \cite{Manning2008Introduction}. The {\em entropy} $\mathrm{H}$ of a set of labeled documents $D$ measures impurity as follows:
$\mathrm{H}(D)= -p^{+}\log_2(p^{+})-p^{-}\log_2(p^{-})$, where $p^{+}$ and $p^{-}$ are the proportions of positive and negative documents in $D$, respectively.\footnote{Note that $\log_2(0)$ is taken to be $0$.}  The information gain $\mathrm{IG}$ of a word $w$ in training set $D$, then,   is the expected entropy reduction due to segmenting on $w$:  $\mathrm{IG}(w, D) = \mathrm{H}(D) - \frac{|D^{w}|}{|D|}\mathrm{H}(D^{w}) - \frac{|\overline{D^{w}}|}{|D|}\mathrm{H}(\overline{D^{w}}),$ where $D^{w}$ is the set of documents in $D$ containing word $w$.  Thus, words with the highest information gain in a training set are expected to be the  most discriminative.  (Although the {\em Mention Facet} can be used for many purposes, populating the facet in this fashion, in a sense, transforms it into yet another kind of {\em Topic Facet}.)   We have also used \appname to populate the {\em Mention Facet} directly from only the positive training documents.  Finally, we supplement this list of discriminative terms with a set of markings for sensitive documents (\eg ``For Official Use Only'', ``FOUO'').


\section{Format, Location, Time, \\and Author Facets}
\label{sec:other}
Our final set of facets are populated through direct extraction from document metadata.  The {\em Format Facet} is populated by tagging documents based on file type (\eg pdf, doc, ppt, txt) and is labeled ``Top File Types.''   The {\em Location Facet} (labeled ``Top Folders'') is populated by tagging each document with the directories in its file path. The {\em Time Facet} (labeled ``Date'' in our application) is populated by extracting the {\em Last-Modified} time from documents.  Finally, the {\em Author Facet} is populated using the {\em Last-Author} or {\em Author} name (when available).  The {\em Location Facet} is displayed as a menu listing the most populous folders, and the {\em Time Facet} is displayed as a calendar widget.  All other facets are displayed as tag clouds. (Note that none of these facets are viewable in Figure \ref{fig:appscreenshot}.) 


\section{Case Studies}
We conduct a series of case studies at the deployment sites using a prototype of our application undergoing field testing.  Motivated by the recent position paper ``Machine Learning That Matters'' by Wagstaff \cite{Wagstaff2012Machine}, we focus on {\em external} validation of our application by assessing time saved and insights gained in collaboration with domain experts. Although our system can be used for many purposes, we focus our evaluation on the current application of interest to our sponsors --- locating information pertaining to military critical technologies within heterogeneous document collections. To locate such information, analysts at the sponsoring agency currently use simple keyword searches exclusively.  Thus, we compare our new approaches to this existing approach.  Since our system employs the use of multiple approaches to locate and discover information, we also draw comparisons {\em among} our new approaches.  For reasons of sensitivity, we cannot reveal the deployment sites, the sponsoring agency, or technical subjects of interest to the agency.  Thus, we redact information as necessary.   

\subsection{Case Study 1:  Search}
\label{sec:eval.search}
\begin{figure}[htb]
\begin{center}
\centerline{\includegraphics[scale=0.25]{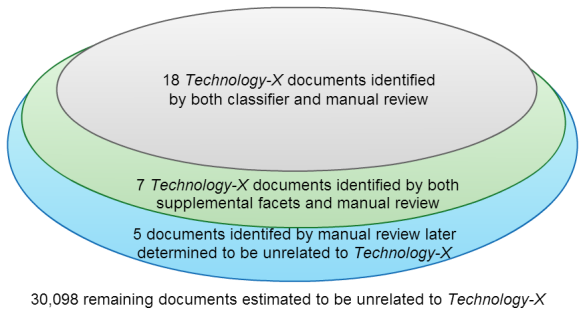}}
\caption{{\scriptsize Venn diagram of search results.  The top (innermost) oval shows documents identified by classifier.  The middle oval shows documents identified by other facets.  The bottom (outermost) oval shows documents identified via a manual review by two analysts.}}
\label{fig:searchresults}
\end{center}
\vskip -0.2in
\end{figure}

Search here involves the task of finding information pertaining to a particular military critical technology within a document collection.  We consider a particular technology of high interest to our sponsors and assess how well the {\em supervised} approaches in our application are able to locate this critical information.  We refer to this technology simply as {\em Technology-X}.  A case was provided to us containing 30,128 files acquired from workstation hard drives of roughly 11 users. The files spanned numerous file formats including Microsoft Office, HTML, PDF, and plain text.   Analysts confirmed to us that the case was positive.  That is, it was manually verified previously to contain information about {\em Technology-X} but not searched thoroughly.  The files were spread across multiple media (\eg external USB hard drives, SATA drives, DVDs). We built machine learning classifiers and a custom mention search  for {\em Technology-X}, as described in Sections \ref{sec:topic.classifier} and \ref{sec:mention}.  Upon loading and indexing the case into our application, we evaluated these approaches and compared results to those obtained from a manual review of the case by two analysts using their existing methodology (\ie {\em ad hoc} keyword searches only).  Results are shown in Figure \ref{fig:searchresults} as a Venn diagram.

As shown, both the classifier and the two analysts identified 18 documents as pertaining to {\em Technology-X}.  The analysts also identified twelve additional files.  Upon review by subject matter expert (SME), only seven of the twelve files were related to {\em Technology-X}, whereas the classifier achieved perfect precision.  Of these seven false negatives, one was a figure with no accompanying text and a second was a 5 sentence email that was deemed critical by the SME.  The most striking result, however, is the time savings achieved.  The two analysts took roughly 7 hours (or 14 person-hours) to locate {\em Technology-X} documents.  By contrast, the classifier identified 18 of the 25 {\em in mere seconds}.  The remaining files (\ie all the seven false negatives) were located in {\em less than 30 minutes} using the {\em Mention Search}, {\em Report  Type Filter}, and {\em Top Folders} facets in our application.  We attribute most false negatives committed by the classifier to the fact that, due to political complications, the positive examples available to us were limited (only 51 examples were used).  Given this and the breadth and depth of military critical technology information, {\em unsupervised} topic discovery is of high importance to this domain.  We discuss this next.

\subsection{Case Study 2:  Discovery}
\label{sec:eval.discovery}

Discovery involves {\em browsing} document collections and allows users to locate information for which they did not even know to look.  A framework to facilitate discovery can clearly facilitate a {\em search} for something specific, as well. Due to logistical and policy-related issues, we were not able to evaluate discovery on the case described in Section \ref{sec:eval.search}.  Instead, we were provided a new case to evaluate, which contained 39,515 files.  Unlike the case from Section \ref{sec:eval.search}, we did {\em not} have any approximation of ground truth, as the case had not been formally reviewed.  Here, we assess the knowledge discovered and summarize lessons learned from execution of our application on this case.

~\\
\noindent
{\bf Identified Critical Topics.}  Table \ref{tab:discoveryresults} shows the two topics pertaining to military critical technologies discovered by our application (referred to as {\em Technology-Y} and {\em Technology-Z}.  Of course, numerous non-critical topics within the document collection were also discovered (some of which were of a personal, non-work-related nature).  As for the critical topics, there were 89 documents found pertaining to {\em Technology-Z} and 232 documents pertaining to {\em Technology-Y} (including duplicate files).  Through a subsequent exhaustive manual review of the case, we estimate that no additional information on military critical technologies of interest was present on this case.  Using our facet-based system, most documents for these two critical topics were identified in less than an hour (and in some cases only minutes). By contrast, domain experts informed us that cases of this size typically require hours or days of analysis to produce similar results, which is consistent with our experience during the manual review.  Several facets were identified as highly effective in identifying these topics when used in combination with each other.  We discuss these next.

\begin{table}[thb]
\centering
{\scriptsize
\begin{tabular}{|l|l|l|} \hline
{\bf Topic} &  {\bf Docs} & {\bf Facets Employed}\\ \hline\hline
\multirow{2}{*}{{\em Technology-Y}} & \multirow{2}{*}{232} &  {\bf Method A:}  Topic Clusters (LDA)$\rightarrow$\appname  \\
&& {\bf Method B:} Report Type Filter~~~$\rightarrow$\appname \\ \hline
\multirow{3}{*}{{\em Technology-Z}} & \multirow{3}{*}{89} & {\bf Method A:}  Topic Clusters (LDA)$\rightarrow$\appname  \\
&& {\bf Method B:}  Report Type Filter~~~$\rightarrow$\appname \\ 
&& {\bf Method C:}  Top Folders~~~~~~~~~~~~~$\rightarrow$\appname \\ \hline
\end{tabular}
\caption{{\scriptsize Critical topics found and effective usage patterns.}}
\label{tab:discoveryresults}
}
\vskip -0.2in
\end{table}

~\\
\noindent
{\bf Effective Usage Patterns Discovered.}  The third column of Table \ref{tab:discoveryresults} displays the facet combinations that were found to be most effective in identifying critical documents.  The combination labeled as {\bf Method A} indicates that the ``Topic Cluster'' facet populated by LDA was first explored and used to filter the search results.  The ``Top Discovered Keywords'' facet (populated by \appname), then, was used to identify document sets related to the critical technology. The combination labeled as {\bf Method B} indicates that the ``Report Type  Filter'' was used to locate documents pertaining to {\em Technical Information} followed again by the ``Top Discovered Keywords'' facet.  Finally, for {\bf Method C}, the ``Top Folders'' facet (\ie our {\em Location Facet}) was used to filter the search results, with the \appname-populated tag cloud being used to quickly assess documents within folders.  Notice that {\em multiple} facet combinations often exist to locate the same set of critical documents.  This highlights a significant advantage to our multi-faceted system:  topics are more likely to be discovered by users when more paths lead towards them.  As shown in the table, {\bf Method A} and {\bf Method B} were employed heavily to find both {\em Technology-Y} and {\em Technology-Z}.  {\bf Method C} was only used for {\em Technology-Z}, not {\em Technology-Y}.  Since the files for {\em Technology-Y} were scattered across many directories, the ``Top Folders'' facet was not as useful.  As can be seen, the ``Top Discovered Keywords'' facet populated by \appname played an indispensable role in all three methods, as it allowed for quick exploration and assessment of the document collection.\footnote{In more recent tests, the {\em Mention Facet}, when used to locate documents with sensitive markings, also was found to be useful in combination with all methods listed in Table \ref{tab:discoveryresults}.} It was particularly useful as a complement to the ``Topic Cluster'' facet, as we now explain.

\begin{figure}
  \centering
  \subfloat[\appname-generated Tag Cloud]{\label{fig:clustlab.kera}\includegraphics[width=0.5\textwidth]{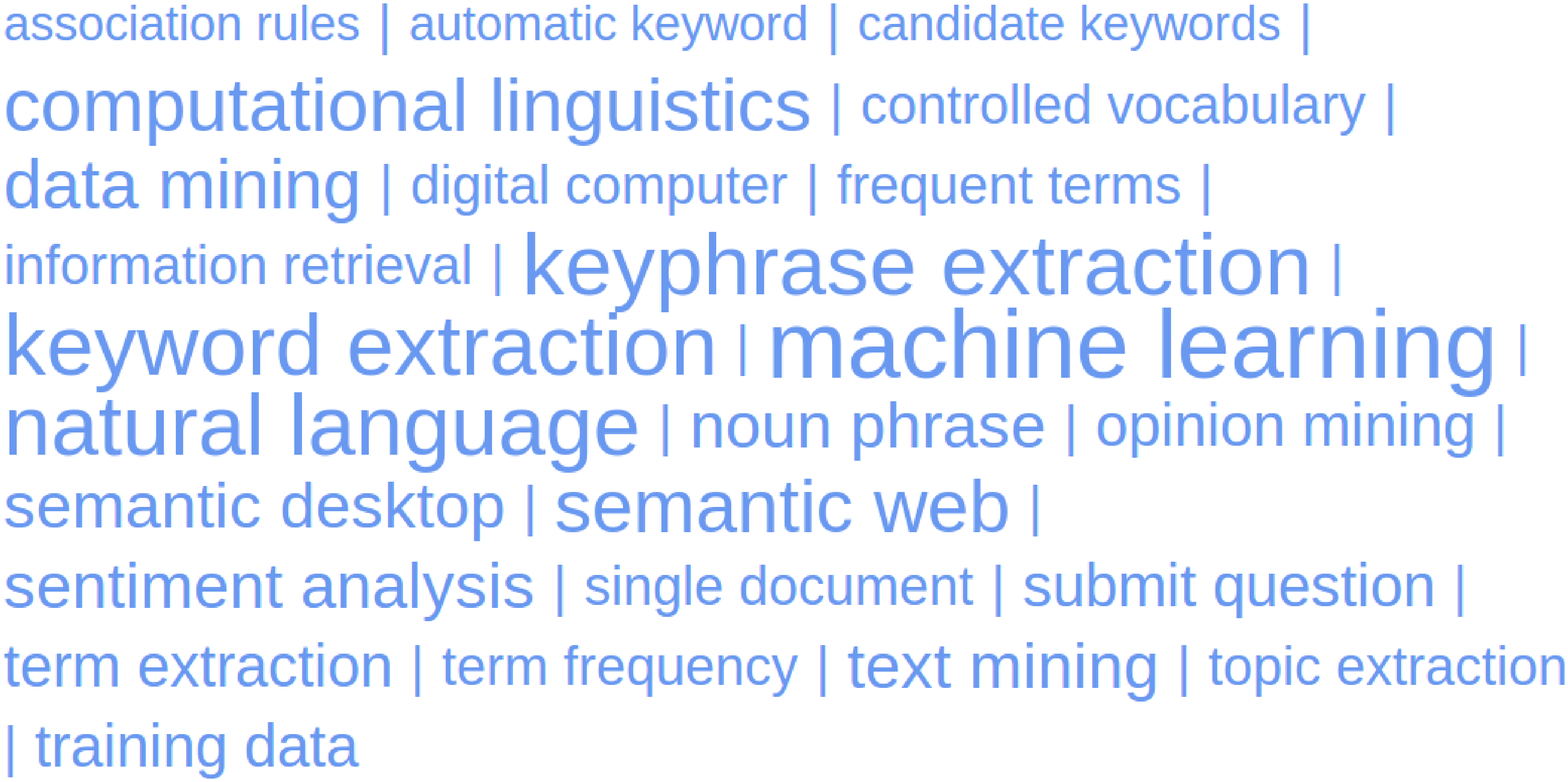}} \\\vspace{.05cm}
    
  \subfloat[Top LDA-Derived Tags] {\label{fig:clustlab.lda}\includegraphics[width=0.5\textwidth]{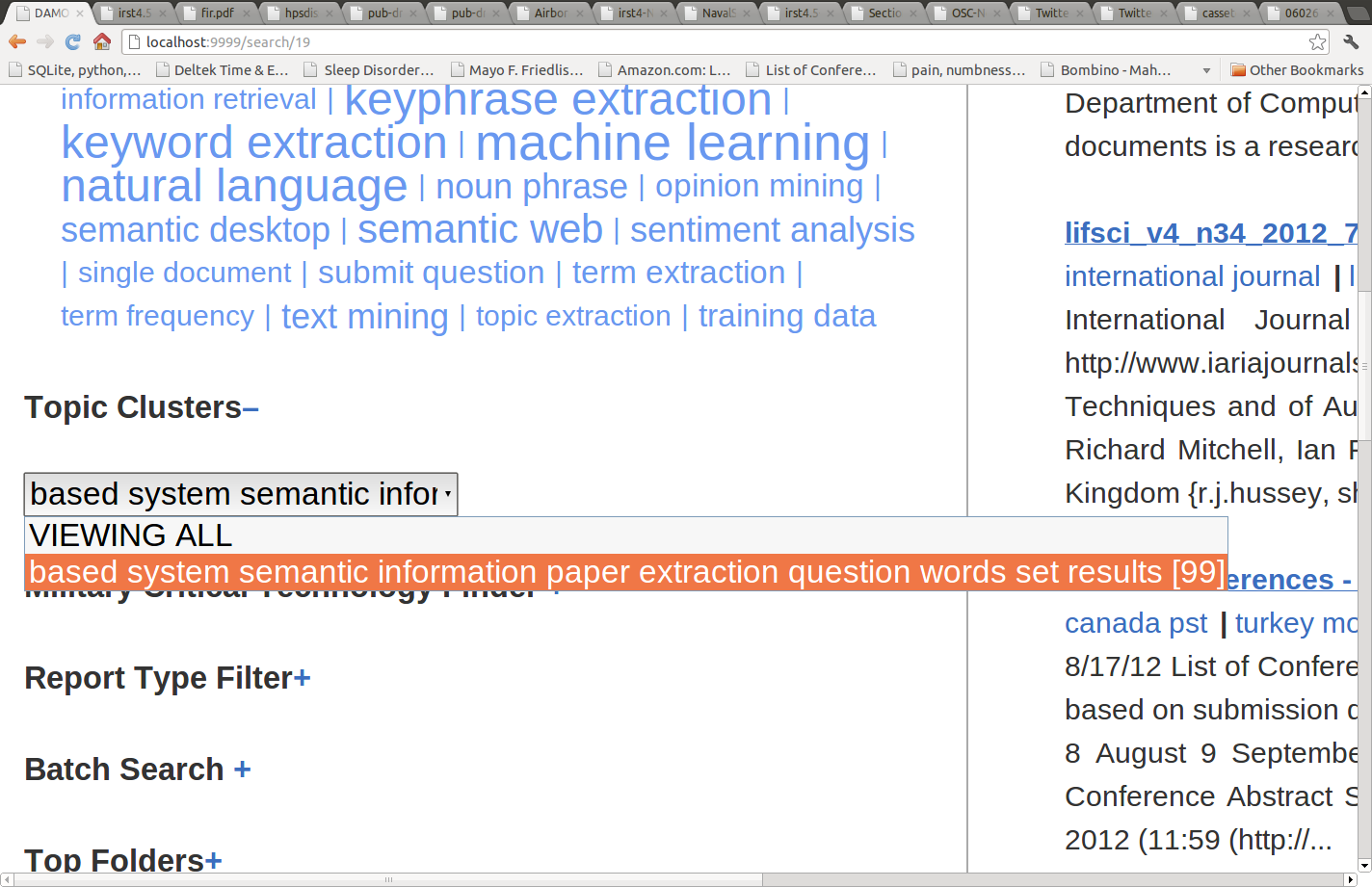}}
\caption{{\scriptsize Tags produced for an LDA-generated cluster of 99 documents about {\em text analytics}:  (a) \appname-generated tag cloud (b) top-ranked tags derived from a typical LDA implementation.  \appname-generated tags tend to be significantly more expressive of the underlying topic, thereby, facilitating exploratory analysis.}}
  \label{fig:clustlab}
  \vskip -0.15in
\end{figure}

~\\
\noindent
{\bf Using \appname as a Cluster Labeling Strategy.}  One of the issues with topic models like LDA is that the terms (or tags) they assign to topics are often not very expressive of the topic.  In other words, in practice, it is quite difficult for humans to go directly from LDA-derived tags to a thematic label for the cluster {\em without} reading documents in the cluster.  This has been recognized in other works on deployed applications based on topic models (\eg see \cite{Wei2010TIARA}), and we found this to be the case in our evaluation, as well.  However, from the effective usage patterns observed previously, we observed that the ``Top Discovered Keywords'' facet populated by \appname is a highly effective way to quickly determine the overall subject matter of a topic cluster (or even a folder).  We cannot illustrate this on topics related to military critical technologies due to their sensitive nature.  Figure \ref{fig:clustlab}, however, shows tags produced for a non-critical document cluster (extracted from documents residing on the first author's laptop).  Although the tag cloud generated by \appname is not quite a thematic label for the cluster, it is significantly more expressive than the tags assigned by a typical LDA implementation.  This, then, illustrates yet another way in which tag cloud facets are useful ``lenses'' into document collections, as described in Section \ref{sec:appoverview}.

~\\
\noindent
{\bf LDA Performance on Critical Technologies.}  Table \ref{tab:ldaperformance} shows the precision and recall with respect to the LDA clustering.  A true positive is defined as placement into an appropriately labeled cluster having a majority of the documents pertaining to the same militarily critical topic (as judged by a SME).  Critical documents placed into clusters with largely non-critical and possibly unrelated documents are considered false negatives.  Non-critical documents appearing in a cluster of largely critical documents are considered false positives. Note the low precision for {\em Technology-Z} resulting from 84 of the 89 critical documents being placed into a larger cluster of non-critical documents.  Since these non-critical documents were indirectly related to the topic covered  by {\em Technology-Z}, LDA was unable to distinguish them from the truly critical documents.  All 89 of these critical documents, however, were ultimately located with help from other facets such as {\em Top Folders} and {\em Top Discovered Keywords}.  These results (combined with an intolerance to false negatives by users in this domain) justify our decision to employ {\em multiple} facets --- as opposed to relying only on topic models, which some other works have done (\eg \cite{Wei2010TIARA}).

\begin{table}[thb]
\centering
{\footnotesize
\begin{tabular}{l|c|c|c} \hline \hline
{\bf Critical Technology}         &  {\bf Precision}    & {\bf Recall}        & {\bf F-Score}   \\ \hline
 {\em Technology-Y}               &   0.89             & 0.98              & 0.93             \\
 {\em Technology-Z}             &   {\bf 0.56}             & 0.94              & 0.70            \\
  \hline \hline
\end{tabular}
\caption{{\scriptsize LDA performance with respect to military critical technologies. Low precision (in bold) and general intolerance to false negatives in this domain motivate our use of multiple information facets.}}
\label{tab:ldaperformance}
}
\vskip -0.2in
\end{table}

~\\
\noindent
{\bf Issues Requiring Future Investigation.}
We conclude our discussion of this case study by noting two issues observed during our evaluation.  The first relates to setting the number of topics, $K$, in LDA.  Most works, including ours, set this value in a largely {\em ad hoc} fashion.  Although there are heuristics and rules-of-thumb that have been proposed (\eg \cite{Can1990Concepts}), most machine learning practitioners acknowledge that the choice of $K$ is ``more art than science.''\footnote{\url{http://cwiki.apache.org/MAHOUT}}  Guessing the correct value of $K$ is particularly difficult for heterogeneous document collections, as $K$ can be severely underestimated.  Moreover, an incorrect setting of $K$ can have detrimental effects on the results.  We have personally found this to be true in our evaluations.  One approach to addressing this is to employ the use of newer non-parametric topic models like HDP {\cite{Teh2006Hierarchical}.  We plan to explore such methods in the future to address these issues.   A second issue relates to \appname.  Although bigrams are appropriate and well-suited for automated tag cloud generation, in some cases, they can produce sub-optimal results (\eg extracting ``Dirichlet allocation'' and not ``latent Dirichlet allocation'').  Some recent approaches to word segmentation based on probabilistic models can potentially be exploited for better keyterm extraction \cite{Goldwater2009Bayesian}.  This, then, is another area for potential future exploration.  

\section{Related Work}
Given the diverse set of facets employed by our application, several different lines of related work exist.  We briefly describe these areas here.

\noindent
{\bf Characterizing Large Document Collections.}  There are several works describing text analytic systems designed to characterized large text corpora (\eg \cite{Kumar2012Navigating,Cselle2007BuzzTrack,Wei2010TIARA}).  Most systems focus on a particular document type (\eg tweets, emails), whereas as our system is designed with heterogeneous document collections in mind.

\noindent
{\bf Topic Modeling, Clustering, and Categorization.}  Many text analytic systems perform topic analysis through use of topic models (\eg LDA \cite{Blei2003Latent}, HDP \cite{Teh2006Hierarchical}) or clustering algorithms like K-Means \cite{Hastie2003Elements,Blei2003Latent}, which are both unsupervised.  Supervised text classification approaches are also sometimes employed \cite{Manning2008Introduction}.  Our objective in this work is to bring to bear {\em multiple} approaches for topic analysis.  As we have shown, using multiple approaches in concert with each other through a faceted browsing framework yields significant advantages.

\noindent
{\bf Keyphrase Extraction.}  Several works describe algorithms to extract keywords and keyphrases from documents.  Some approaches are supervised or require an entire corpus as input (\eg \cite{Witten1999KEA,Bun2002Topic}), which, as described previously, is not suitable for our purposes.  TextRank \cite{Mihalcea2004TextRank} and RAKE \cite{Rose2010Automatic} are two approaches that are purely unsupervised and operate on individual documents. However, as we have shown, they do not appear well-suited to automated tag cloud generation.  A related area of research is collocation extraction (\eg \cite{Manning1999Foundations,Dunning1993Accurate,Pecina2005Extensive}), which we exploit in  the \appname algorithm.

\noindent
{\bf Tag Cloud Research.}  Numerous works leverage tag clouds for both faceted navigation and corpus visualization (\eg \cite{Kuo2007Tag,Zubiaga2009ContentBased,Knautz2010Tag}). The overwhelming majority of this work focuses on manually-generated tags (\eg social-tagging systems) as opposed to automated generation of tags, which is one of the foci of our work.

\section{Conclusion}

In this paper, we have proposed a demonstrably effective system for exploratory analysis of arbitrary document collections.  Our system, based on multiple {\em information facets}, is designed to address a major capability gap within the U.S. federal government:  investigative analysis of highly heterogeneous document collections.   We have presented a concrete implementation of this multi-faceted system that aids users in identifying information pertaining to military critical technologies embedded within large and arbitrary document collections. A prototype of our application was successfully deployed in May 2013.  In the future, we plan to extend the tool in numerous ways including sentence-based summaries of topics and visualizations of topic clusters.

\balancecolumns
\end{document}